\documentclass[conference]{IEEEtran}
\IEEEoverridecommandlockouts    

\usepackage{xcolor}
\usepackage{lipsum}
\usepackage{amssymb}
\usepackage{graphicx}
\graphicspath{{./Fig/}}
\usepackage{graphics} 
\usepackage{algpseudocode}
\usepackage{cite}
\usepackage{amsfonts}
\usepackage{graphicx}
\usepackage{epstopdf}
\usepackage{subfigure}
\usepackage{textcomp}
\usepackage{color}
\usepackage{multirow}
\usepackage{makecell}
\usepackage{algorithm}
\usepackage{booktabs}
\usepackage{multirow}
\usepackage{bbm}
\usepackage{diagbox}
\usepackage{soul} 
\usepackage{amsmath} 
\usepackage{enumitem}
\usepackage[letterpaper, left=0.75in, right=0.75in, bottom=0.75in, top=1in]{geometry}

\title{\LARGE \bf Learning from Human Driving: A Human-in-the-Loop Online Behavior Cloning Framework for Autonomous Driving}

\author{Yuhong Shi$^{1}$, Jianyi Liu$^{1,*}$, Lihang Sun$^{1}$, Li Li$^{1}$, Xudong Dong$^{1}$
\thanks{$^{1}$ Yuhong Shi, Jianyi Liu, Lihang Sun, Li Li, and Xudong Dong are with the State Key Laboratory of Human-Machine Hybrid Augmented Intelligence, Institute of Artificial Intelligence and Robotics, Xi'an Jiaotong University, Xi’an 710049, China.}
\thanks{This work was supported in part by the National Key Research and Development Program of China under grant No. 2024YFB3309303, and in part by the Xi'an Jiaotong University-China Mobile Communications Group Co., Ltd. Joint Institute. }
\thanks{$*$ Corresponding author: Jianyi Liu (Email: {\tt\small jyliu@xjtu.edu.cn})}
}

\begin{document}
\maketitle
\thispagestyle{empty}
\pagestyle{empty}
	
\begin{abstract}
\textcolor{black}{
With the evolution of large foundation models (LFMs), data-driven autonomous driving has made significant strides. However, existing paradigms still face severe challenges in complex interaction and long-tail scenarios due to distribution shift and causal confusion. These limitations often result in a lack of human-level decision-making flexibility and safety in extreme conditions.
To overcome this limitation, this paper proposes a Human-in-the-Loop Online Behavior Cloning framework (HiL-OBC) for autonomous driving, which aims to deeply integrate the cross-modal perceptual capabilities of LFMs with the high-level driving intelligence of human experts. Specifically, HiL-OBC deployment is executed through three critical phases: policy initialization with human intervention, latent behavioral modeling with Bayesian policy adaptation, and online deployment and updates. Furthermore, we design a Multi-modal Online Behavior Cloning (MOBC) model, which optimizes the base driving policy online through a lightweight network architecture, a takeover trigger mechanism, and a multi-variant loss function, thereby enhancing the system's decision-making robustness in complex environments.
We evaluated the HiL-OBC on the LangAuto-Human CARLA benchmark. Experimental results demonstrate that the driving policies optimized via the human-in-the-loop mechanism achieve substantial performance gains: the DS of StructNav, LFG, and LMDrive increased by 47.25\%, 31.59\%, and 32.12\%, respectively, with a simultaneous reduction in traffic violations. Additionally, a detailed analysis of various experimental settings and key components highlights the advantages of human-in-the-loop learning in improving decision-making robustness and overall driving performance.
}
\end{abstract}


\section{INTRODUCTION}

Autonomous driving has entered a transformative era with the advent of large foundation models (LFMs), which endow agents with unprecedented semantic depth and cross-modal perception \cite{10657019}. Despite strong semantic skills, these models suffer from a major split. High-level cognitive reasoning and fine-grained motor execution remain disconnected, making it hard to translate abstract logic into precise driving action \cite{driving}.

We contend that bridging this gap necessitates a paradigm shift from large-scale data scaling to proactive policy alignment \cite{10682977}. The infinite variety of real-world traffic cannot be fully captured by any static, offline dataset \cite{shift}. Instead of relying on more data, we advocate for dynamic online calibration, where the model learns and adapts during actual deployment. This motivates our Human-in-the-Loop Online Behavior Cloning (HiL-OBC) framework. 
Unlike traditional systems that treat human drivers as a last-resort backup, HiL-OBC treats human intervention as an authoritative supervisory signal that actively reshapes the model’s internal logic in real-time. By situating human expertise directly within the decision-making loop, HiL-OBC transforms expert intuition into persistent corrective signals to refine the model’s decision boundaries and resolve the compounding errors that plague data-driven policies.

The strength of HiL-OBC lies in its ability to internalize human intelligence rather than just mimicking it. We replace rigid ``all-or-nothing" takeover rules with a probabilistic arbitration mechanism for fluid authority transition. Central to this approach is the Multi-modal Online Behavior Cloning (MOBC) model, which generates a takeover probability $P_{to}$ that functions as a differentiable gating mechanism. Rather than a discrete switch, $P_{to}$ enables a weighted fusion between the base agent’s generalized knowledge and the expert’s localized corrections. This formulation ensures a smooth policy shift, allowing the system to adaptively leverage human guidance where the base policy's confidence falters. Most critically, we formulate the policy evolution as a Bayesian Posterior Adaptation. By treating the policy parameters as a dynamic distribution, HiL-OBC performs variational inference to assimilate new intervention data effectively evolving the policy while explicitly quantifying the system's epistemic uncertainty.

In summary, we propose a holistic HiL-OBC framework as in Fig. \ref{fig1} that integrates the driving intelligence of LFMs with the high-order intelligence of human experts, further enhancing the performance of base driving policies. Our main contributions include:

\begin{figure*}[htbp]
  \centering
  \includegraphics[width=0.8\linewidth]{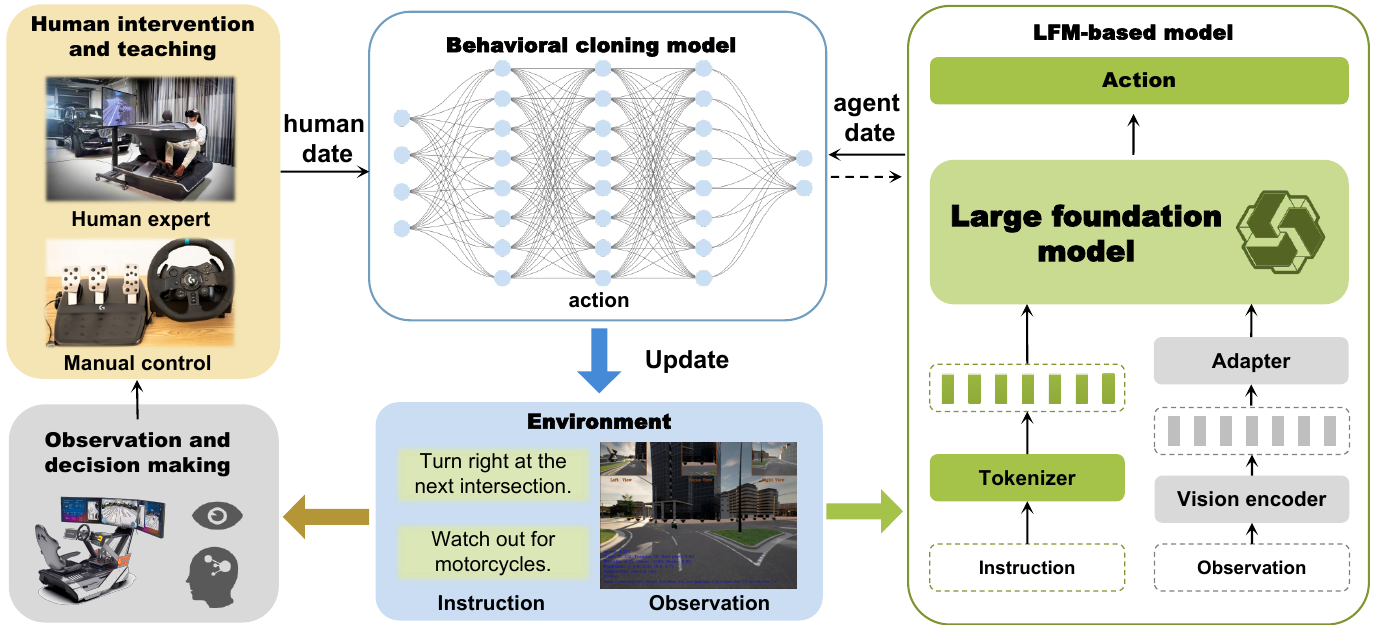}
  \caption{
A human-in-the-loop online behavior cloning framework for autonomous driving.
  }  \label{fig1}
\vspace{-1em}
\end{figure*}

\begin{itemize}
\item We propose a HiL-OBC framework for autonomous driving, which integrates initial policy pretraining and intervention data collection, latent behavioral modeling with Bayesian policy adaptation, and online deployment with closed-loop updates into a unified framework.
\item We design a MOBC model that achieves online optimization of the basic policy through a lightweight model structure, a takeover triggering mechanism, and a multivariate loss function.
\item We construct the LangAuto-Human CARLA dataset to evaluate the performance of a closed-loop driving system incorporating human driving actions.
\item We conduct extensive experiments to validate the effectiveness and superiority of the HiL-OBC and analyze the impact of different experimental settings and components on the framework performance.
\end{itemize}


\section{Related Works}
\subsection{Large Foundation Model-based Autonomous Driving}

Large foundation models have introduced a novel research paradigm in autonomous driving by leveraging their advanced contextual reasoning and multimodal data understanding capabilities \cite{10682977}. 
Based on LLMs, DriveGPT4 \cite{10629039} processes multi-frame video inputs and textual queries to generate vehicle action descriptions, reasoning for actions, and low-level control signals. Graph-based Visual Question Answering (VQA) techniques have been employed to enhance visual understanding. ReasonDrive \cite{ReasonDrive} and SimpleLLM4AD \cite{zheng2024simplellm4adendtoendvisionlanguagemodel} leverage VQA to represent objects and their relationships within a scene, enabling more precise scene analysis and decision-making, while effectively addressing the long-tail challenges in complex traffic scenarios.
Additionally, VLMs offer a deeper understanding of visual information. DriveVLM \cite{DriveVLM} aligns VLMs with autonomous driving systems and behavior planning states, utilizing Chain-of-Thought reasoning modules to achieve scene description, scene analysis, and hierarchical planning. DriveVLM-Dual further enhances spatial reasoning and real-time planning capabilities. 
Despite these advancements, relying on large foundation models for navigation scene understanding faces challenges in accurately modeling the driving process. The generated driving actions often lack human-like driving behaviors and habits, as well as effective strategies for handling long-tail problems.

\subsection{Learning from Human Intervention }

HiL is an emerging technological paradigm that synergizes human intelligence with machine intelligence to enhance decision-making, safety, and adaptability through real-time human feedback and intervention. In recent years, researchers in the field of autonomous driving have explored various approaches to incorporating human intelligence.
Within the reinforcement learning (RL) framework, \cite{10507015} proposes a human-guided deep reinforcement learning framework that optimizes autonomous driving decisions through real-time feedback from human experts. Focusing on safety, \cite{10596046} introduces a shared control framework that integrates human drivers with autonomous driving systems to improve robustness. \cite{HIL1} examines the role of HiL in optimizing driving strategies from the perspectives of expert guidance and experience-prioritized learning, respectively.
In decision-making and planning research, \cite{9761167} adopts a penalty-factor-based cooperative control method and a reference-free RL framework to regulate steering torque and obstacle avoidance. \cite{10400976} explores a hybrid augmented intelligence framework that integrates human intelligence with evolutionary algorithms. From a brain-inspired modeling perspective, \cite{10127583} investigates the potential of HiL in achieving human-like driving decisions within mixed-traffic environments. 
The integration of human intelligence through HiL has proven to be a transformative approach to enhancing the decision-making, safety, and adaptability of autonomous driving systems.
	

\begin{figure*}[htbp]
  \centering
  \includegraphics[width=0.8\linewidth]{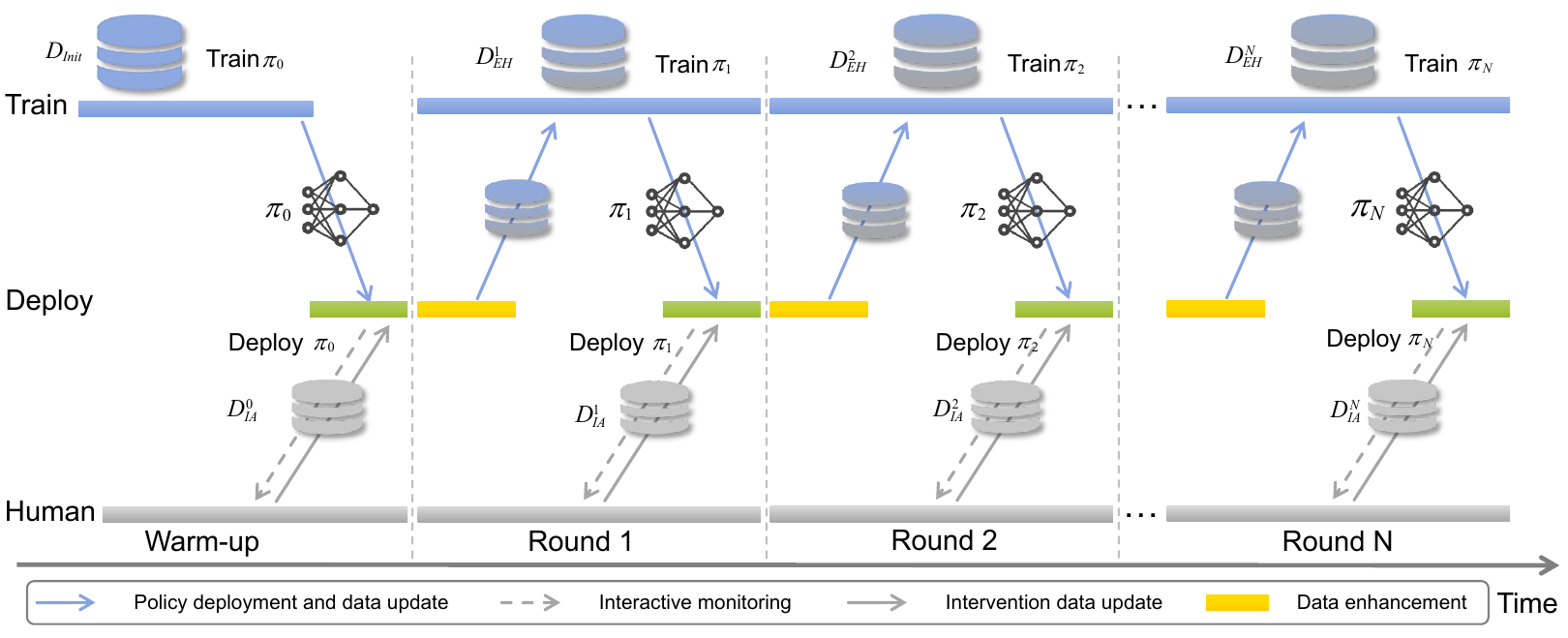}
  \caption{The deployment workflow of the proposed HiL-OBC.
Policy deployment, model training, and human intervention occur simultaneously and in parallel. After the initial model training, it is deployed on the robot to acquire human intervention data for the next round of training.}  \label{figpd}
\vspace{-1em}
\end{figure*}

\section{Methodology}
\label{sec:Methodology}

\subsection{Problem Formulation}

We formulate the HiL-OBC framework as a sequential decision-making problem with a gating mechanism. 
Our main focus is on framework deployment, takeover assessment, and online policy optimization in dynamic environments.
We model the driving process as a multimodal Markov Decision Process defined by 
\begin{equation}
\mathcal{M} = (\mathcal{S}, \mathcal{A}, \mathcal{P}),
\end{equation}
where $\mathcal{S} = \mathcal{X}_{vis} \times \mathcal{X}_{mot}$ represents the state space, $\mathcal{X}_{vis} = \{g_l, g_m, g_r\}$ denotes visual observations captured by left, middle, and right cameras, and $\mathcal{X}_{mot}$ represents the ego vehicle's motion state vector $state_{motion}  \in \mathbb{R}^d$, includes velocity, position, acceleration, heading angle, and current control signals.
Within HiL-OBC, the base agent generates control actions ${a_{base}^{t}} \in {\cal A}$, while a multimodal online behavioral cloning model predicts the takeover probability $P_{to} \in [0, 1]$. This probability serves as the basis for triggering human expert intervention and supervision. Our method does not rely on static training but emphasizes online learning from human corrections, minimizing the gap between the agent's policy and expert behavior through policy optimization.

\subsection{Framework Deployment}

The proposed HiL-OBC lies in processing human-intervention data and achieving seamless model deployment and continuous online adaptation. The framework's evolutionary mechanism integrates Bayesian inference and dynamic parameter distribution optimization, aiming to efficiently and robustly internalize corrective knowledge provided by human experts into the model's policy.

\subsubsection{Policy Initialization and Human Intervention Acquisition}

The system undergoes pre-training utilizing the initial dataset $\mathcal{D}_{Init}$ from a simulation environment, obtaining a base policy $\pi_0$ endowed with fundamental navigation capabilities. Subsequently, $\pi_0$ is deployed within the interactive simulator CARLA \cite{dosovitskiy2017carla}. During the deployment phase, human experts monitor the agent's decision-making and execution in real-time via a supervisory interface. 
Upon detecting risky behaviors or path deviations, experts intervene through two modalities: 1) Prompting Takeover, where high-level commands are issued; 2) Manual Intervention, where experts directly manipulate the controller to output a sequence of optimal actions $\tau = \{a_{human}^t\}$. The system automatically logs environmental states $s_t$, original model outputs $a_{base}^t$, expert corrective actions $a_{human}^t$, and intervention types, forming a high-quality set of corrective pairs $(s_t, a_{base}^t, a_{human}^t)$. These intervention data $\mathcal{D}_{IA}$ are incrementally integrated into the initial dataset via the DAgger \cite{ross2011reduction} to construct an incremental dataset $\mathcal{D}_{EH}$ as in Fig. \ref{figpd}. Fundamentally, this dataset encapsulates the distribution shift between the model's current policy and the expert's policy within high-risk, long-tail scenarios.

\subsubsection{Latent behavioral modeling and Bayesian policy adaptation}

After obtaining the $\mathcal{D}_{{EH}}$, the system does not perform a full retraining. Instead, it initiates a parameter posterior update process grounded in variational inference. Specifically, we introduce a conditional variational autoencoder (VAE) as a latent model of expert behavior. Conditioned on the environment state $s_t$, the VAE learns to encode the expert-corrected action $a_{human}^t$ into a low-dimensional and smooth latent space $z_t$, from which the action can be reconstructed. Training on $\mathcal{D}_{\text{EH}}$ enables the VAE to capture fine-grained action patterns and stylistic characteristics of expert behavior under critical states.

More importantly, we treat the parameters $\theta_t$ of the policy network as random variables and impose a prior distribution $p(\theta_t)$ over them. New evidence is incorporated via Bayesian updating to compute the posterior distribution.
\begin{equation}
p(\theta_t \mid \mathcal{D}_{{EH}}) \propto p(\mathcal{D}_{{EH}} \mid \theta_t)p(\theta_t),
\end{equation}
where the likelihood term $p(\mathcal{D}_{{EH}} \mid \theta_t)$ is evaluated using the VAE model, which measures the similarity between actions generated under the current policy parameters  $\theta_t$ and the expert-corrected actions. We adopt stochastic gradient variational Bayes to approximately infer this posterior, yielding an updated parameter distribution $q_\phi(\theta_{t})$. The mean vector of this distribution is taken as the optimized policy parameter $\theta_{t+1} $.
In essence, this approach leverages human intervention data to softly refine the probability distribution over the entire parameter space, leading to more stable updates while explicitly quantifying model uncertainty.

\subsubsection{Online deployment and update}

The above learning procedure is designed as a lightweight, periodically triggered online fine-tuning model that operates in parallel with continuous system deployment. As illustrated in Fig. \ref{figpd}, the workflow as follows:
\begin{enumerate}[label=(\arabic*)]
  \item The model is executed online under the current policy $\pi_t $, while intervention data are collected through a combination of active DAgger \cite{ross2011reduction} querying and passive logging.
  \item When the data buffer reaches a predefined capacity or when the model’s performance metrics indicate a bottleneck, the online learning model is activated.
  \item The model leverages the most recent intervention dataset $\mathcal{D}_{{IA}} $ to perform representation alignment via the VAE and to carry out a fast Bayesian posterior update of the policy parameters.
  \item The updated policy $\pi_{t+1} $ is seamlessly substituted for the online model through a Warm-update mechanism, after which the system immediately continues interaction and data collection based on the improved policy.
\end{enumerate}

The complete process is summarized in Algorithm \ref{ALG_Online_BC}.
This mechanism ensures that the system can continuously and efficiently transform the local corrections made by human experts into robust improvements to the global policy distribution of the model, ultimately achieving a gradual alignment of the agent's behavior distribution with the expert distribution without frequent intervention.

\begin{algorithm}[ht]
    \caption{MOBC}
    \label{ALG_Online_BC}
    \renewcommand{\algorithmicrequire}{\textbf{Input:}}
    \renewcommand{\algorithmicensure}{\textbf{Output:}}
    \begin{algorithmic}[1]
     \Statex  \algorithmicrequire Initial dataset LangAuto-Human $\mathcal{D}_{Init}$, base policy $\pi_{base}$, learning rate $lr$, number of epochs $\tau$, network parameters $\pi_{\theta}$  
    \Statex  \algorithmicensure  Optimized policy $\pi_{opt}$    
        \State Initialize policy $\pi_{0} \gets \pi_{base}$
        \State Initialize dataset $\mathcal{D}_{Init} \gets \text{LangAuto-Human}$  
        \For{$i = 1$ to $\tau$} 
            \State Deploy policy $\pi_{0}$ with human intervention
            \State Collect new corrective pairs $(s_t, a_{base}^t, a_{human}^t)$
            \State Add new data $(s_t, a_{base}^t, a_{human}^t)$ to dataset $\mathcal{D}_{EH}$: 
            $\mathcal{D}_{EH} \gets \mathcal{D}_{Init} \cup (s_t, a_{base}^t, a_{human}^t)$
            \State Train MOBC with $\mathcal{D}_{EH}$ to update $\pi_{\theta,i}$
            \State Generate new policy $\pi_{\theta,i}$ using MOBC
            \If{$\pi_{\theta,i}$ performs better than $\pi_\text{opt}$}
                \State Update $\pi_{opt}$: $\pi_{opt}\gets \pi_{\theta,i}$
            \EndIf
        \EndFor
        
        \State \Return Optimized policy $\pi_{opt}$\;
    \end{algorithmic}
\end{algorithm}

\subsection{Multi-modal Online Behavior Cloning}
\subsubsection{Network Structure}
\begin{figure}[hbp]
  \centering
  \includegraphics[width=0.45\textwidth]{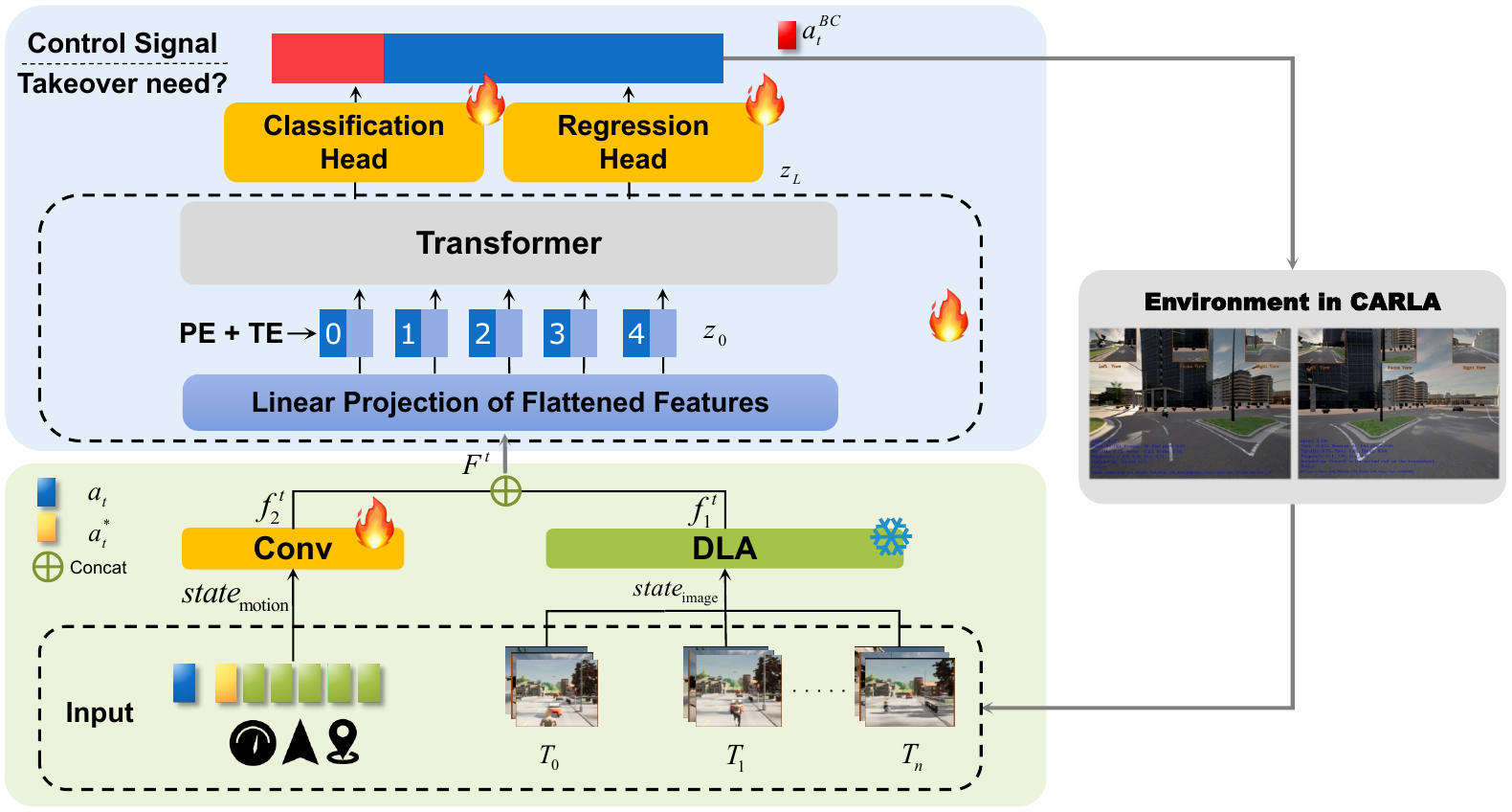}
  \caption{\textcolor{black}{The structure of the MOBC, which consists of two major components: 1) a state encoder, which processes multi-view images and driving states for scene understanding and generates visual representations; 2) a 4-layer Transformer backbone with a regression prediction head, which predicts control signals and takeover probabilities. ``PE'' and ``TE'' represent pose embedding and temporal embedding, respectively.}
  }
  \label{fig:ViT-BC}
\end{figure}

For robot policy, we constructed a tiny MOBC for decision-making as in Fig. \ref{fig:ViT-BC}. The driving state is jointly represented by vehicle motion information and multi-view images. The motion state includes velocity, position, acceleration, heading angle, and current control signals, while the visual state consists of left, center, and right camera views, all acquired from vehicle-mounted sensors.
In the visual branch, a DLA-34 backbone \cite{yu2018deep} is employed to extract multi-scale features. The motion branch encodes temporal dynamics using convolutional neural networks. After dimensional transformation and fusion, the two branches are integrated into a unified state representation $F^t$. Based on this, we model the temporal correlation of actions.
The sequence input contains a learnable action token. Following processing through multiple Transformer layers, a global feature representation is obtained and used to regress steering, throttle, and braking commands. In parallel, a classification head is retained to predict the takeover probability, enabling joint modeling of action regression and risk assessment.

\subsubsection{Takeover Trigger}

In the proposed HiL-OBC, the intervention of human experts is not random or continuous, but controlled by a hybrid triggering logic based on real-time risk assessment and performance monitoring.
During the data acquisition and model deployment intervention phases, takeovers are implemented via rule-based triggers. We have designed a multivariate conditional function $T(s_t, a_{base}^{t}, \hat{s}_t)$. When this function exceeds a predefined threshold, the system initiates the takeover process, allowing human expert actions to override the current autonomous commands. These expert action data are subsequently integrated into the dataset for downstream policy learning. 
Specifically, the multivariate conditional function considers distance to other traffic participants, deviation from the desired path, driving safety, and comfort, formulated as
\begin{equation}
\begin{aligned}
T(s_t, a_{base}^{t}, \hat{s}_t) =& \;  \omega_1 \cdot \mathrm{ReLU}(d_{\min} - d) + \omega_2 \cdot |e_{\text{path}}| \\
&+ \omega_3 \cdot \big(1 - {cos}(s_t, \hat{s}_t)\big),
\end{aligned}
\end{equation}
where $s_t$ and $\hat{s}_t$ denote the current and predicted environmental state, $ a_{base}^{t}$ represents the action planned by the model, $d_{min}$ and $d$ are the distances from the desired trajectory and the current trajectory to the nearest obstacle, $e_{path}$ signifies the path-tracking error, $cos(s_t, \hat{s}_t)$ represents the cosine similarity between the current driving state and the desired state, and $\omega_1, \omega_2, \omega_3$ are set to 0.2, 0.4, 0.4, respectively.

During the inference and online optimization phase, we transition from rigid rule-based switching to a probabilistic arbitration mechanism. 
To ensure a smooth transfer of control authority, the final executed policy $\pi_{opt}(s_t)$  is formulated as a gated fusion between the agent's policy $\pi_{base}$ and the learned human policy $\pi_{human}$:
\begin{equation}
\pi_{opt}(s_t) = P_{to}(s_t) \cdot \pi_{human}(s_t) + (1 - P_{to}(s_t)) \cdot \pi_{base}(s_t)
\end{equation}
where $P_{to}(s_t)$ is obtained by training the model on the current dataset and represents the confidence level of whether human intervention should be carried out.

\subsubsection{Loss Function}

The primary objective of MOBC is to learn a parameterized policy $\pi_{\theta}(s)$ that generates the actions $ a_{base}^{t}$ to closely approximate the expert's actions $ a_{human}^{t}$ in a particular state $s$. 
To bridge the gap between autonomous execution and human intervention, we formulate a joint optimization objective that simultaneously minimizes the action regression error and the takeover classification loss. The integrated loss function is defined as
\begin{equation}
\scalebox{0.8}{
$\begin{aligned}
\theta^* = \arg \min_{\theta} \; 
\mathbb{E}_{(s, a^*) \sim \mathcal{D}} \Big[
& \gamma_1  \mathcal{L}_{act}(\pi_{\theta}(s), a_{human}^t) + \gamma_2 \mathcal{L}_{brk}(\pi_{\theta}^{brk}, y_{brk}) \\
& + \gamma_3 \mathcal{L}_{to}(\pi_{\theta}^{p}(s), \mathbb{I}_{human})
\Big].
\end{aligned}$
}
\end{equation}
where $\mathbb{E}_{(s,a^*) \sim \mathcal{D}}$ denotes the empirical expectation on the dataset $\mathcal{D}$, and $\gamma_1, \gamma_2, \gamma_3$ represent the coefficients assigned to balance different tasks, $y_{{brk}} \in \{0, 1\}$ denotes the binary label for the braking extracted from expert demonstrations, and $\mathbb{I}_{human}$ represents the indicator function $T(s_t, a_{base}^{t}, \hat{s}_t)$.
Specifically, we employ mean square error as $\mathcal{L}_{act}$ to ensure precise maneuver imitation. For the discrete control signals and safety-critical assessments, we utilize Focal Loss \cite{lin2017focal} for both the braking classification $\mathcal{L}_{brk}$ and the takeover probability prediction $\mathcal{L}_{to}$. 
In our implementation, we prioritize risk assessment and steering precision by setting $\gamma_1 = 1.0$, $\gamma_2 = 0.8$, and $\gamma_3 = 1.2$. This scheme ensures that the model remains sensitive to potential takeover scenarios while maintaining smooth longitudinal and lateral control.

\subsection{Human Intervention and Data Aggregation}
\label{sec: Imitate human intelligence}
\subsubsection{LangAuto-Human Dataset}
\begin{figure}[htbp]
  \centering
  \includegraphics[width=0.9\linewidth]{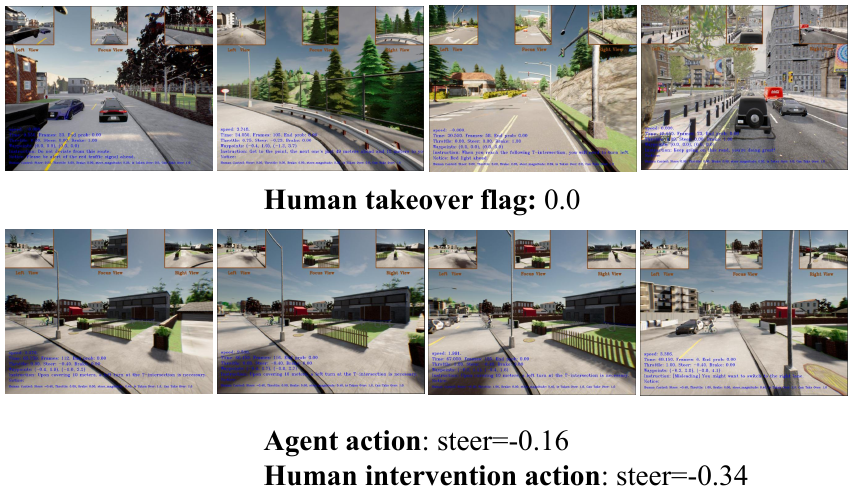}
  \caption{
  Two examples of the collected data with corresponding
  Human intervention flags and optimal human actions.
  }  \label{fig:dataset}
\end{figure}

We constructed the LangAuto-Human (Language-guided Autonomous driving with Human-in-the-loop) dataset in CARLA \cite{dosovitskiy2017carla}, designed to train models with human intervention. Compared to the LangAuto dataset, we have incorporated human intervention and recorded human actions and state data. 
Specifically, the LangAuto-Human dataset covers 8 publicly available towns in the CARLA simulator \cite{dosovitskiy2017carla} and various scenarios (e.g., highways, intersections, and roundabouts). We also consider 16 types of environmental conditions, including combinations of 7 weather conditions (Clear, Cloudy, Wet, MidRain, WetCloudy, HardRain, SoftRain) and 3 daylight conditions (Night, Noon, Sunset). The LangAuto-Human dataset across 32 long trajectories, 16 short trajectories, and 16 tiny trajectories of the LangAuto benchmark. Human drivers intervene and apply actions through the keyboard when the agent makes an error, and the human actions and corresponding states are recorded. The comparison before and after human intervention is depicted in Fig. \ref{fig:dataset}. Each frame with intervention information includes the corresponding human intervention flag and the optimal human action.

\subsubsection{Data Augmentation}
Since the base system performs well in most straight-driving scenarios, human intervention is only required at intersections and during turns. This leads to a scarcity of data involving human intervention. To address the issue of sample imbalance, we employ data augmentation techniques. By flipping and rotating the camera images captured during human intervention, we increase the data volume to 200K frames.

\begin{table*}[htbp]
\centering
\caption{Performance comparisons with various models on the LangAuto benchmark. We report metrics from three evaluation runs.}
\setlength{\tabcolsep}{3mm}
\label{tab1}
\resizebox{0.88\textwidth}{!}{
\begin{tabular}{ccccccccccc}
\toprule
\multirow{2}{*}{Method} & \multicolumn{3}{c}{LangAuto} & \multicolumn{3}{c}{LangAuto-Short} & \multicolumn{3}{c}{LangAuto-Tiny} \\ \cmidrule(r){2-4} \cmidrule(r){5-7} \cmidrule(r){8-10}
 & DS$\uparrow$ & RC$\uparrow$ & IS$\uparrow$ & DS$\uparrow$ & RC$\uparrow$ & IS$\uparrow$ & DS$\uparrow$ & RC$\uparrow$ & IS$\uparrow$ \\ 
\midrule
Random & 1.783 & 4.928 & 0.129 & 2.031 & 5.542 & 0.223 & 2.561 & 7.461 & 0.246 \\
Greedy & 3.251 & 13.250 & 0.229 & 10.351 & 16.512 & 0.315 & 14.012 & 19.245 & 0.414 \\
EnvDrop \cite{EnvDrop} & 15.258 & 23.414 & 0.568 & 23.920 & 36.124 & 0.615 & 28.475 & 39.272 & 0.653 \\
VLN-BERT \cite{VLBERT} & 17.532 & 26.001 & 0.617 & 28.731 & 39.155 & 0.734 & 34.027 & 48.541 & 0.701 \\
HAMT \cite{HAMT} & 17.601 & 25.311 & 0.613 & 28.951 & 38.091 & 0.722 & 34.914 & 47.198 & 0.699 \\
StructNav \cite{SNav} & 20.458 & 31.674 & 0.698 & 35.892 & 46.124 & 0.715 & 48.745 & 59.832 & 0.752 \\
LFG \cite{LFG} & 24.235 & 38.882 & 0.673 & 42.168 & 55.452 & 0.732 & 48.892 & 56.328 & 0.774 \\
LMDrive \cite{10657019} & 26.955 & 46.910 & 0.724 & 50.410 & 61.221 & 0.745 & 52.084 & 64.712 & 0.787 \\ 
\midrule
StructNav+MOBC & 30.125 & 49.874 & 0.625 & 59.458 & 68.124 & 0.824 & 68.459 & 74.125 & 0.805 \\
LFG+MOBC & 31.892 & 46.874 & 0.642 & 46.124 & 55.236 & 0.841 & 55.236 & 61.459 & 0.879 \\
LMDrive+MOBC & \textbf{35.613} & \textbf{54.789} & \textbf{0.650} & \textbf{64.036} & \textbf{73.091} & \textbf{0.856} & \textbf{72.589} & \textbf{79.924} & \textbf{0.920} \\ 
\bottomrule
\end{tabular}}
\end{table*}

\begin{table*}[htbp]
\centering
\setlength{\tabcolsep}{3.5mm}
\caption{Performance comparison of experienced and inexperienced human on the LangAuto benchmark. We report the metrics for 3 evaluation runs.}
\resizebox{0.88\textwidth}{!}{
\begin{tabular}{cccccccccc}
\toprule
\multirow{2}{*}{Experience}& \multicolumn{3}{c}{LangAuto} & \multicolumn{3}{c}{LangAuto-Short} & \multicolumn{3}{c}{LangAuto-Tiny} \\ \cmidrule(r){2-4} \cmidrule(r){5-7} \cmidrule(r){8-10}
                        &  DS $\uparrow$  & RC $\uparrow$   &  IS $\uparrow$      &  DS $\uparrow$  & RC $\uparrow$   &  IS $\uparrow$  &  DS $\uparrow$  & RC $\uparrow$   &  IS $\uparrow$   \\ \cmidrule(r){1-1} \cmidrule(r){2-4} \cmidrule(r){5-7} \cmidrule(r){8-10} 
                  Experienced    &  \textbf{35.613}  &  \textbf{54.789}  &  \textbf{0.650}  &   \textbf{64.036}   &  73.091    &  \textbf{0.856}   &  \textbf{72.589}    &  79.924   &  \textbf{0.920}    \\
                  Inexperienced      &  34.499  &  54.031  &  0.639  &   62.819   &  \textbf{74.150}    &  0.848 &  71.065    &  \textbf{80.516}   &  0.873    \\
                        \bottomrule
\end{tabular}}
\label{table:qualification}\vspace{-1em}
\end{table*}

\begin{table*}[htbp]
\centering
\setlength{\tabcolsep}{3mm}
\caption{Performance comparison of human intervention duration on the LangAuto benchmark. We report the metrics for 3 evaluation runs.}
\resizebox{0.88\textwidth}{!}{
\begin{tabular}{cccccccccc}
\toprule
\multirow{2}{*}{Human intervention Duration} & \multicolumn{3}{c}{LangAuto} & \multicolumn{3}{c}{LangAuto-Short} & \multicolumn{3}{c}{LangAuto-Tiny} \\ \cmidrule(r){2-4} \cmidrule(r){5-7} \cmidrule(r){8-10}
                        &  DS $\uparrow$  & RC $\uparrow$   &  IS $\uparrow$      &  DS $\uparrow$  & RC $\uparrow$   &  IS $\uparrow$  &  DS $\uparrow$  & RC $\uparrow$   &  IS $\uparrow$   \\ \cmidrule(r){1-1} \cmidrule(r){2-4} \cmidrule(r){5-7} \cmidrule(r){8-10} 
                0      &  26.955  &  46.910  &  0.724  &    50.410   &  61.221   &  0.845   &   52.084    &  64.712    &  0.787    \\
                1500      &  26.452  &  35.129  &  \textbf{0.753}  &      48.568   &  68.714    &  0.726  &  60.142    &    80.028   &  0.792    \\
                3000      &  30.683  &  47.358  &  0.649  &    53.885   &  68.521    &  0.783  &    69.258    &  \textbf{81.209}   &  0.852    \\
                4500      &  33.925  &  52.591  &  0.638  &    55.792  &  70.863    &  0.787  &    70.194    &  76.429   &  0.918    \\
                All    &  \textbf{35.613}  &  \textbf{54.789}  &  0.650  &   \textbf{64.036}   &  \textbf{73.091}    &  \textbf{0.856}  &  \textbf{72.589}    &  79.924   &  \textbf{0.920}    \\
                        \bottomrule
\end{tabular}}
\label{table:duration}\vspace{-1em}
\end{table*}

\begin{table*}[htbp]
\centering
\setlength{\tabcolsep}{3 mm}
\caption{Performance comparison of 4 backbones on the LangAuto benchmark. We report the metrics for 3 evaluation runs.}
\resizebox{0.88\textwidth}{!}{
\begin{tabular}{cccccccccc}
\toprule
\multirow{2}{*}{Feature Backbone} & \multicolumn{3}{c}{LangAuto} & \multicolumn{3}{c}{LangAuto-Short} & \multicolumn{3}{c}{LangAuto-Tiny} \\ \cmidrule(r){2-4} \cmidrule(r){5-7} \cmidrule(r){8-10}
                        &  DS $\uparrow$  & RC $\uparrow$   &  IS $\uparrow$      &  DS $\uparrow$  & RC $\uparrow$   &  IS $\uparrow$  &  DS $\uparrow$  & RC $\uparrow$   &  IS $\uparrow$   \\ \cmidrule(r){1-1} \cmidrule(r){2-4} \cmidrule(r){5-7} \cmidrule(r){8-10} 
                  ResNet-18 \cite{he2016deep}      &  28.478  &  49.600  &  0.576  &   
                          52.095 &   64.031  &  0.805   &  
                          58.672 &   66.974  &  0.846    \\
                  ResNet-50 \cite{he2016deep}      &  28.831  &  50.012  &  0.581  
                  &   53.104   &  66.723    &  0.812  
                  &  60.859    &  70.541   &  0.863    \\
                  Hourglass-104 \cite{newell2016stacked}      &  30.634  &  44.223  &  \textbf{0.718}  
                  &   56.036   &  70.892    &  0.793  
                  &  63.589    &  77.189   &  0.824    \\
                  DLA-34 \cite{yu2018deep}     &  \textbf{35.613}  &  \textbf{54.789}  &  0.650  
                  &   \textbf{64.036}   &  \textbf{73.091}    &  \textbf{0.836}  &  \textbf{72.589}    &  \textbf{79.924}   &  \textbf{0.920}    \\
                        \bottomrule
\end{tabular}}
\label{table:backbone}
\end{table*}

\section{Experiments}

\subsection{Experiment Settings and Details}
We implement and evaluate our approach on the CARLA simulator of version 0.9.10.1 \cite{dosovitskiy2017carla}. All experiments are performed on an Ubuntu 20.04 system equipped with two NVIDIA RTX 3090 GPUs and a 2.0 GHz Intel Xeon Gold 6330 CPU.

\textbf{Metrics.}
We selected three main metrics: Route Completion Rate (RC), Violation Score (IS), and Driving Score (DS) to evaluate the agent's driving performance as in \cite{10657019}. Among these, DS is computed as the product of RC and IS, which is usually regarded as the main evaluation indicator.
Additionally, we adopt Driving Accuracy (DA) as a measure of training performance. If the model correctly predicts both the required intervention and the corresponding actions, it is considered a successful expert operation. DA is defined as the percentage of correctly predicted expert operations over the total number of operations within the entire trajectory.

\textbf{Training.}
After data augmentation, the ratio of samples including human intervention to samples where the agent operates without intervention is approximately 1:4. The learning rate of the MOBC network is set to $7.5 \times 10^{-4}$ and is trained by the AdamW optimizer. The number of training epochs is 125, including 15 warm-up epochs. We trained our model using 2 NVIDIA RTX 3090 GPUs, and the training process took 12 hours to complete.

\begin{figure*}[htpb]
    \centering
    \includegraphics[width=0.85\textwidth]{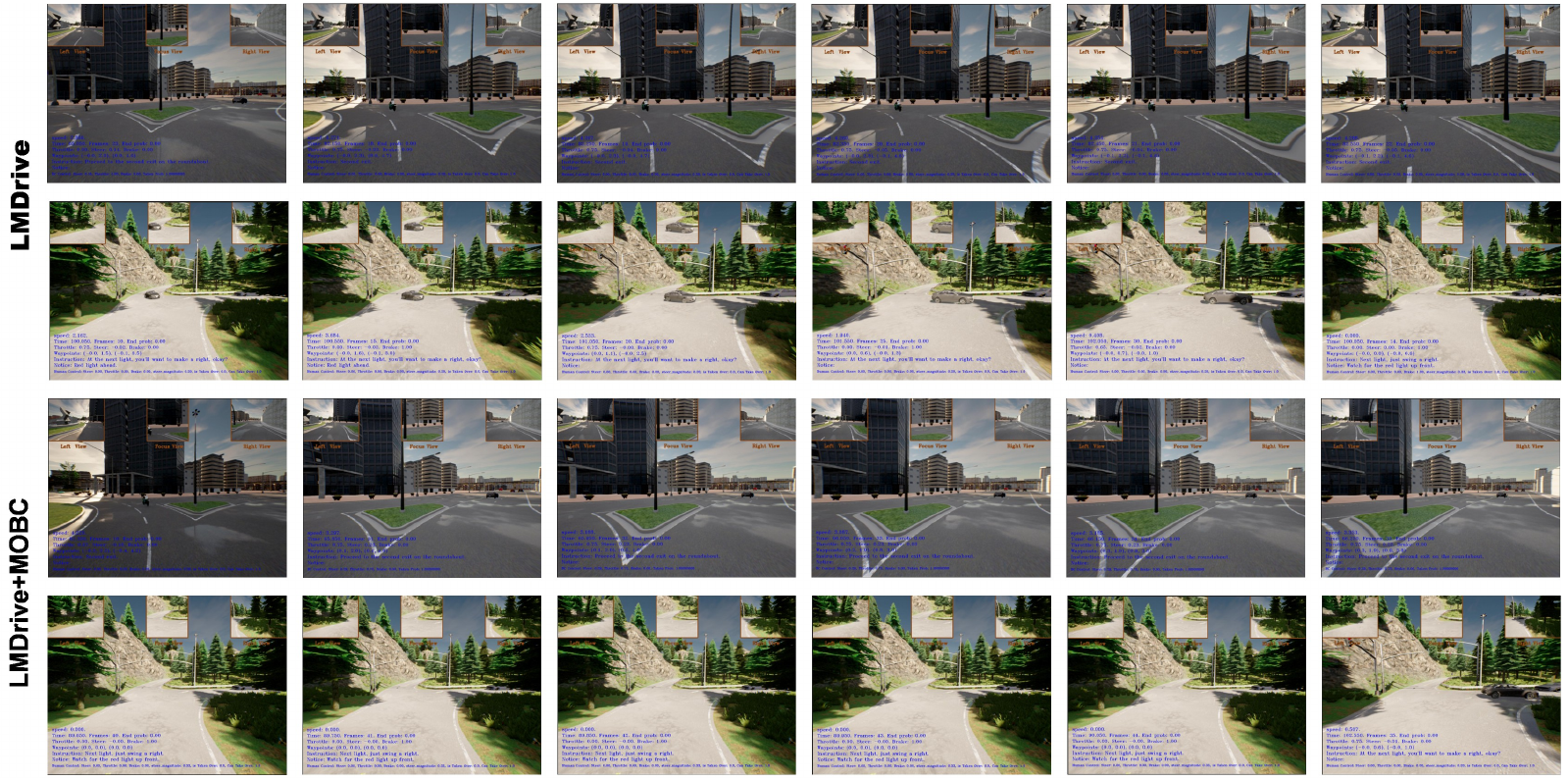} %
    \caption{In the navigation comparison between LMDrive and MOBC, the first trajectory shows the LMDrive agent deviating from the intended path by taking a leftward route. The MOBC successfully corrected this erroneous action, properly executing a right turn. In the second trajectory, the LMDrive agent violated traffic light rules, while MOBC continued to brake until the green light appeared.}
    \label{fig:visualization}
    \vspace{-1em}
\end{figure*}

\subsection{Results Analysis}
\subsubsection{Main Results}
To evaluate the performance of the HiL-OBC, we selected StructNav \cite{SNav}, LFG \cite{LFG}, and LMDrive \cite{10657019} as base policies for validation in identical scenarios. Furthermore, the performance of the models augmented with human data was compared against current SOTA methods. Experiments were primarily conducted on the long, short, and tiny trajectories within the LangAuto dataset.
As synthesized in Table \ref{tab1}, compared with the base policies StructNav \cite{SNav}, LFG \cite{LFG}, and LMDrive \cite{10657019}, HiL-OBC consistently yields varying degrees of improvement in DS in long, short, and micro trajectories. Notably, the largest DS gain is achieved by StructNav \cite{SNav} on short trajectories, reaching 65.66\%.
These results indicate that MOBC effectively leverages human intervention data to substantially enhance model performance. Additionally, MOBC is capable of extracting traffic light information from visual inputs, thus mitigating traffic violations and improving the IS. As illustrated in Fig. \ref{fig:visualization}, after optimization with MOBC, LMDrive \cite{10657019} not only corrects behaviors that lead to error trajectories, but also prevents actions that would result in traffic violations.

Furthermore, compared with SOTA models, the LMDrive+MOBC model achieves the best performance across all evaluation metrics and trajectories. This can be attributed to the strong base policy and the greater performance gains enabled by human-in-the-loop optimization.

\subsubsection{Ablation Study}

\textbf{Experience and Proficiency.} 
To investigate the impact of human experts' driving experience and proficiency on model optimization performance, we recruited human experts with varying levels of driving proficiency to intervene in the driving process.
Based on driving experience, the human experts were divided into two groups and intervened under the same conditions, resulting in two different human expert intervention datasets.
The results in Table \ref{table:qualification} show that human experience has less impact on performance. Compared to inexperienced drivers, experienced drivers improved DS performance by 3.1\%, 1.9\%, and 2.1\%. 
This can be attributed to the fact that the driving tasks in the simulator are relatively simple for human experts, and most participants can correct the obvious trajectory deviations caused by LMDrive \cite{10657019}. Such interventions do not require fine-grained operations that heavily rely on prior driving experience.

\textbf{Duration of Human Intervention.} In Table \ref{table:duration}, we investigated the impact of the duration of human intervention on performance during driving. We set the maximum number of frames for human intervention to 0, 1500, 3000, 4500, and All, where 0 indicates without human intervention, and ``All'' indicates human intervention in the whole process. The results show that human intervention within 1500 frames leads to a significant decrease in performance, potentially even underperforming the baseline LMDrive \cite{10657019}. A possible reason for this is that the average human intervention time in the dataset falls between 2000 and 3000 frames, and too little human data may lead to incorrect predictions of the takeover probability, resulting in unknown actions.

\begin{table}[htbp]
\centering
\setlength{\tabcolsep}{4.5mm}
\caption{Comparison of Ablation Study Results on Individual Components.}
\resizebox{0.96\linewidth}{!}{
\begin{tabular}{cccc}
\toprule
Module design  &  DS $\uparrow$  & RC $\uparrow$   &  IS $\uparrow$       \\ \cmidrule(r){1-1} \cmidrule(r){2-4} 
                  
                  w/o State Encoder     & 32.696   &  44.622  & 0.605\\ 
                  w/o TF     & 25.230   &  36.986  & 0.576  \\ 
                  w/o DAgger \cite{ross2011reduction}     & 12.663   &  13.747  & 0.370  \\ 
                   MOBC (Full)       &  \textbf{35.613}  &  \textbf{54.789}  &  \textbf{0.650} \\
                                         \bottomrule
\end{tabular}}
\vspace{-0.5mm}
\vspace{-1em}                                   
\label{table:ablation}
\end{table}

\begin{table}[htbp]
\centering
\setlength{\tabcolsep}{2mm}
\caption{Training Loss and Driving Accuracy in Training Progress}
\resizebox{0.96\linewidth}{!}{
\begin{tabular}{cccccc}
\toprule
\diagbox{Epochs}{Loss}  &  $\mathcal{L}$   & $\mathcal{L}_\text{act}$   &  $\mathcal{L}_\text{brk}$    & $\mathcal{L}_\text{to}$  &  DA   \\ \cmidrule(r){1-1} \cmidrule(r){2-6} 
                  0     & 1.644 & 0.159 & 0.814 & 0.670 & 26.61\\
                  25    & 0.579 & 0.085 & 0.212 & 0.281 & 45.19\\
                  50    &  0.232 & 0.025 & 0.086 & 0.121 & 50.71\\
                  75    & 0.228 & 0.019 & 0.085 & 0.121 & 60.97\\
                  100   & 0.226 & 0.021 & 0.085 & 0.120 & 58.73\\
                  125   & 0.223 & 0.018 & 0.084 & 0.120 & 53.27\\
                        \bottomrule
\end{tabular}}
\label{table:loss}\vspace{-1em}
\end{table}

\textbf{Feature Backbone} 
We evaluate our method using four different image feature extraction backbones: ResNet18 \cite{he2016deep}, ResNet50 \cite{he2016deep}, DLA-34 \cite{yu2018deep}, and Hourglass-104 \cite{newell2016stacked}. In Table \ref{table:backbone}, we observe that DLA-34 \cite{yu2018deep} is capable of aggregating semantic information, and significantly outperforms the other backbones. This suggests that understanding semantic information, such as traffic lights and obstacles in view, is crucial for action prediction.

\textbf{Each Component} 
 In Table \ref{table:ablation}, we conduct ablation studies on each component of the HiL-OBC. 
 Firstly, we removed the State Encoder (denoted as ``w/o State Encoder") and directly fed the image into the decoder with a regression prediction head. The DS decreased to $32.696$ without aggregation of traffic semantic information.  Subsequently, we removed the Transformer layer and utilized FC layers to directly map the hidden dimension to the action dimension (denoted as ``w/o TF"). The DS further decreased to 25.230. The primary reason for the decline is the lack of attention to correlations and positional features when considering only independent state-action pairs. Lastly, we removed the DAgger \cite{ross2011reduction} loop and trained the model offline on a fixed dataset (denoted as ``w/o DAgger \cite{ross2011reduction}"). Due to insufficient data leading to compounding errors, the DS finally dropped to 12.663.

\textbf{Loss} 
In Table \ref{table:loss}, we present training loss and DA at every 25 epochs. Overall, the loss decreases rapidly and gradually converges to around 0.22, while the DA steadily increases, reaching its peak at epoch 75. The $\mathcal{L}_\text{act}$ and $\mathcal{L}_\text{brk}$ decrease swiftly, indicating a fast learning rate of the model. However, $\mathcal{L}_\text{to}$ decreases more slowly due to the sparsity of the data.

\section{Conclusion}

In this paper, we introduced a human-in-the-loop online behavior cloning framework for autonomous driving, aiming to enhance autonomous driving performance. The implementation of this framework is based on the MOBC, which learns expert data through a state encoder, action decoder, and online learning. Additionally, we introduce LangAuto-Human, a dataset for human-in-the-loop autonomous driving. 
Extensive experiments demonstrate that our approach effectively leverages human driving experience to optimize system decision-making in complex driving scenarios while mitigating the compounding error inherent in behavior cloning. 

However, the reliance on the CARLA simulation environment presents certain limitations in terms of sim-to-real generalization. In practical real-world deployment, the framework must contend with significant stochasticity, such as sensor noise that affects state perception and communication latency that may disrupt the synchrony of the human-in-the-loop pipeline. Furthermore, the inherent unpredictability of human behavior in mixed-traffic environments not fully captured in a controlled simulation. Therefore, future research will focus on bridging this reality gap by incorporating uncertainty estimation, robust control mechanisms for latency compensation, to ensure reliability in diverse and noisy real-world conditions.


	\bibliographystyle{IEEEtran}
	\bibliography{root} 
	
\end{document}